\documentclass[runningheads]{llncs}
\usepackage{graphicx}
\usepackage{nicematrix,enumitem}
\usepackage{booktabs}

\begin{document}
\title{Enhancing Clinical Support for Breast Cancer with Deep Learning Models using Synthetic Correlated Diffusion Imaging}

\titlerunning{Leveraging Synthetic Correlated Diffusion Imaging for Breast Cancer}

\author{Chi-en Amy Tai\inst{1}\orcidID{0000-0001-7023-8784}\email{amy.tai@uwaterloo.ca}
\and Hayden Gunraj\inst{1}\orcidID{0000-0002-1028-5797} 
\and Nedim Hodzic\inst{1}
\and Nic Flanagan\inst{1} 
\and Ali Sabri\inst{2,3}\orcidID{0000-0002-4335-1770}
\and Alexander Wong\inst{1,4,5}\orcidID{0000-0002-5295-2797}}
\authorrunning{C. Tai et al., amy.tai@uwaterloo.ca}
\institute{Department of Systems Design Engineering, University of Waterloo, Waterloo, ON, Canada \and Department of Radiology, McMaster University, Hamilton, ON, Canada \and Niagara Health System, St. Catharines, ON, Canada \and Waterloo AI Institute, University of Waterloo, Waterloo, ON, Canada \and DarwinAI Corp., Waterloo, ON, Canada}

\maketitle              

\begin{abstract}
Breast cancer is the second most common type of cancer in women in Canada and the United States, representing over 25\% of all new female cancer cases. As such, there has been immense research and progress on improving screening and clinical support for breast cancer. In this paper, we investigate enhancing clinical support for breast cancer with deep learning models using a newly introduced magnetic resonance imaging (MRI) modality called synthetic correlated diffusion imaging (CDI\textsuperscript{s}). More specifically, we leverage a volumetric convolutional neural network to learn volumetric deep radiomic features from a pre-treatment cohort and construct a predictor based on the learnt features for grade and post-treatment response prediction. As the first study to learn CDI\textsuperscript{s}-centric radiomic sequences within a deep learning perspective for clinical decision support, we evaluated the proposed approach using the ACRIN-6698 study against those learnt using gold-standard imaging modalities. We find that the proposed approach can achieve better performance for both grade and post-treatment response prediction and thus may be a useful tool to aid oncologists in improving recommendation of treatment of patients. Subsequently, the approach to leverage volumetric deep radiomic features for breast cancer can be further extended to other applications of CDI\textsuperscript{s} in the cancer domain to further improve clinical support. 

\end{abstract}

\section{Introduction}
Breast cancer is the second most common type of cancer in women in Canada and the United States, representing over 25\% of all new female cancer cases~\cite{cancer-stats}. As such, there has been immense research and progress on improving screening techniques and processes to proactively detect the presence of breast cancer in individuals at risk~\cite{https://doi.org/10.48550/arxiv.1606.05718}. However, it is estimated that 2,261,419 new cases of breast cancer were diagnosed across the world in 2020~\cite{cancer-stats} and predicted that 43,700 American women will die from breast cancer in 2023~\cite{american-breast-cancer}.

However, not all breast cancer is fatal. When patients are first diagnosed with breast cancer, they are categorized into two main types: in situ and invasive breast cancer~\cite{american-breast-cancer}. The former is a less severe form of breast cancer that is a precursor to the latter type. The latter type, invasive breast cancer, represents approximately 80\% of diagnosed cases and signifies that the cancer has already or can spread into the nearby tissue areas~\cite{american-breast-cancer,cancer-stats}. 

\begin{figure}
    \vspace{-2em}
  \centering
    \includegraphics[width=\linewidth]{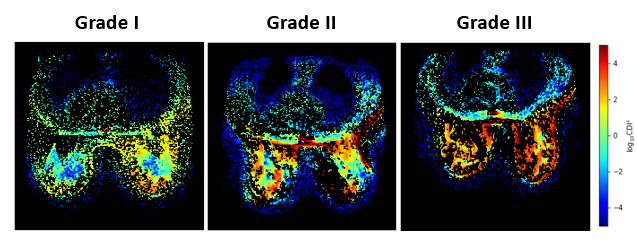}
    \vspace{-2em}
  \caption{Example breast CDI\textsuperscript{s} images for the SBR grades.}
  \vspace{-1em}
  \label{fig:grade-grid}
\end{figure}

Patients with invasive breast cancer also often receive a breast cancer grade that represents the similarity of the cancer cells to normal cells under the microscope. The three breast cancer grades (low, intermediate, and high) describe the speed of growth and likelihood of a good prognosis. Low grade (grade 1) cancer has the best prognosis with slow  growth and spread of the cancer, while high grade (grade 3) cancer has the worst prognosis with the greatest difference between cancer and normal cells and represent cancer that is fast-growing with quick spread to other cells. As such, the stage and grade of breast cancer are vital factors used to determine the severity of breast cancer and discern the best treatment strategy as the stage and grade have been shown to relate to the success of various treatment strategies~\cite{grading-future}. Specifically, the gold-standard Scarff-Bloom-Richardson (SBR) grade (with example CDI\textsuperscript{s} shown in Figure~\ref{fig:grade-grid}) has been shown to consistently indicate a patient's response to chemotherapy~\cite{sbr-useful}. 

Unfortunately, the gold-standard method of grading the breast cancer is currently determined by a pathologist looking at a tissue sample from the cancer tumour under a microscope. As such, the current method to determine the grade requires removal of some cancer cells from the patient which can lead to stress and discomfort along with high medical costs~\cite{american-breast-cancer}.

\begin{figure}
    \tiny
    \centering
    \vspace{-2em}
    \includegraphics[width=\textwidth]{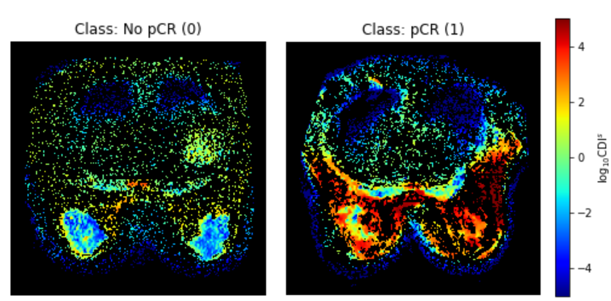}
    \caption{Example breast CDI\textsuperscript{s} images with and without pCR.}
    \label{fig:pCR-ex}
     \vspace{-2em}
\end{figure}

Following grading, surgery is commonly administered to prevent breast cancer from further developing and to remove cancerous tissue~\cite{american-breast-cancer}. However, some non-metastatic breast cancer tumors are inoperable~\cite{survival-bc-patients}. Recently, a type of treatment termed neoadjuvant chemotherapy has risen in usage as it can shrink a large tumor before surgery (so that the tumor can become operable)~\cite{survival-bc-patients} and it may also result in a pathologic complete response (pCR) which is the absence of active cancer cells present in surgery~\cite{pcr-bc}. Example breast CDI\textsuperscript{s} images with and without pCR is shown in Figure~\ref{fig:pCR-ex}. However, neoadjuvant chemotherapy is expensive, time-consuming, and may expose patients to radiation as well as lead to other significant side effects such as reduced fertility~\cite{cost-effectiveness-nac}. 

The current process to recommend neoadjuvant chemotherapy is based on the expert, but human judgment of the medical oncologist and/or radiation oncologist of whether the patient will live longer and benefit from the treatment~\cite{nac-human-recommendation}. With potential biases and high uncertainty in clinical judgment~\cite{human-judgement-problem}, there is potential for some erroneous recommendations leading to some patients later developing preventable detrimental advanced cancer or being exposed to unnecessary radiation. 

For these clinical support tasks for breast cancer, pathologists typically consult the patient's MRI images. Current gold-standard MRI modalities include diffusion-weighted imaging (DWI), apparent diffusion coefficient (ADC), and T2-weighted (T2w). DWI is a form of MRI that measures the motion of water molecules within the tissue with the b values (0, 100, 600, 800) denoting the specific configuration of the scanner such as gradient strength with 0 indicating no diffusion sensitivity and a greater sensitivity as b increases~\cite{MARTINEZHERAS2021490}. ADC is the value obtained by taking the slope of the curve created with the different b values with lower ADC indicating regions with restricted diffusion or potentially cancerous tissue~\cite{dwi-prediction}. T2w is a type of contrast MRI image that enhances water signals~\cite{VM208346402021.72720}. Recently, synthetic correlated diffusion imaging (CDI\textsuperscript{s}) was introduced as a promising imaging modality for clinical decision support for prostate cancer~\cite{prostate-cdis}. CDI\textsuperscript{s} introduces synthetic signals by extrapolating MRI data to introduce more data points by analyzing the direction of diffusion in the cancerous tissue. 

In this paper, we investigate enhancing clinical support for breast cancer with deep learning models using a newly introduced magnetic resonance imaging (MRI) modality called synthetic correlated diffusion imaging (CDI\textsuperscript{s}). More specifically, we leverage a volumetric convolutional neural network to learn volumetric deep radiomic features from a pre-treatment cohort and construct a predictor based on the learnt features for breast cancer SBR grading and post-treatment response prediction to neoadjuvant chemotherapy. The dataset utilized in this study is derived from the American College of Radiology Imaging Network (ACRIN) 6698 study, which is a comprehensive multi-center study aimed at collecting patient medical images along with their treatment response to neoadjuvant breast cancer treatment~\cite{acrin6698-data-1,acrin6698-data-2,acrin6698-data-3,acrin6698-data-4}. As the first study to learn CDI\textsuperscript{s}-centric radiomic sequences within a deep learning perspective for clinical decision support, we evaluated the proposed approach using the ACRIN-6698 study against those learnt using gold-standard imaging modalities.

\section{Related Works}
\subsection{Breast Cancer Grading}
Previous studies have examined the merit of pairing computer vision techniques for breast cancer grade prediction using radiomics~\cite{radiomics}, statistical tests~\cite{dwi-prediction}, elasticity ratios~\cite{elastography}, multitask learning models~\cite{dce}, and deep learning~\cite{image-analysis}. A comprehensive review of radiomics discussed the high potential of tumor grade prediction using radiomics on breast imaging~\cite{radiomics} and Burnside et al. demonstrated that computer-extracted image phenotypes on magnetic resonance imaging (MRI) could accurately predict the breast cancer stage~\cite{comp-tumors}. However, Surov et al. concluded that diffusion-weighted imaging used with the Mann-Whitney U test was inapt at predicting breast cancer tumour grades~\cite{dwi-prediction}. On the other hand, deep learning methods to identify metastatic breast cancer~\cite{https://doi.org/10.48550/arxiv.1606.05718} and breast cancer grade~\cite{image-analysis} have presented high accuracies of over 80\% with a review on invasive breast cancer supporting the importance of leveraging artificial intelligence on grade prediction~\cite{grading-future}.

\subsection{Pathologic Complete Response Prediction}
In the past, a variety of different modalities and methods were investigated to predict pathologic complete response with patient features such as using a nonparametric Mann-Whitney test for diffusion-weighted imaging (DWI) and MRS~\cite{prediction-dwi-mrs}, logistic regression models on MRI images~\cite{biomarkers-predicting}, hard threshold parameter values~\cite{nac-bc-r}, AdaBoost classifier with qCT features~\cite{ml-qCT}, and an assortment of machine learning models with qCT features~\cite{extended-apriori-prediction}. Furthermore, previous studies have also examined the usage of deep learning and volumetric data with breast cancer. Convolutional neural network algorithms were studied to predict post-NAC axillary response with breast MRI images~\cite{predicting-post-neoadjuvant}, a three-layer 3D CNN architecture was trained to detect breast cancer using a dataset of 5547 images with an AUC of 0.85~\cite{3d-cnn-bc-detect}, and convolutional neural networks with 3D MRI images were used to predict pCR to neoadjuvant chemotherapy in breast cancer~\cite{prediction-pcr-3dcnn}.

\section{Methodology}
\subsection{Patient Cohort and Imaging Protocol}
The pre-treatment (T0) patient cohort in the American College of Radiology Imaging Network (ACRIN) 6698/I-SPY2 study was used as the patient cohort in this study~\cite{acrin6698-data-1,acrin6698-data-2,acrin6698-data-3,acrin6698-data-4}. After removing patients that had incomplete data, 252 patient cases remained for use for the task of grading, and 253 patients for the task of pathologic complete response prediction. The ACRIN study contained MRI images across 10 different institutions for patients at four different timepoints in their treatment~\cite{acrin6698-data-1,acrin6698-data-2,acrin6698-data-3,acrin6698-data-4}. However, only the timepoint T0 was used as patients at this stage had not received any neoadjuvant chemotherapy and thus, the images would be most representative of the ones that pathologists would evaluate to determine SBR grade and decide if the patient should receive neoadjuvant chemotherapy. To compare the performance of CDI\textsuperscript{s} with current gold-standard MRI modalities used in clinical practice, diffusion-weighted imaging (DWI) acquisitions, T2-weighted (T2w) acquisitions, and apparent diffusion coefficient (ADC) maps were also obtained. Using the DWI images, we also obtain CDI\textsuperscript{s} acquisitions for each of the patient cases. Finally, the SBR grade for each breast cancer patient and the post-treatment pCR to neoadjuvant chemotherapy was also obtained after pre-treatment imaging was conducted to facilitate for learning and evaluation purposes.

\subsection{Extracting Deep Radiomic Sequences from CDI\textsuperscript{s}}
\begin{figure}
  \centering
    \includegraphics[width=\textwidth]{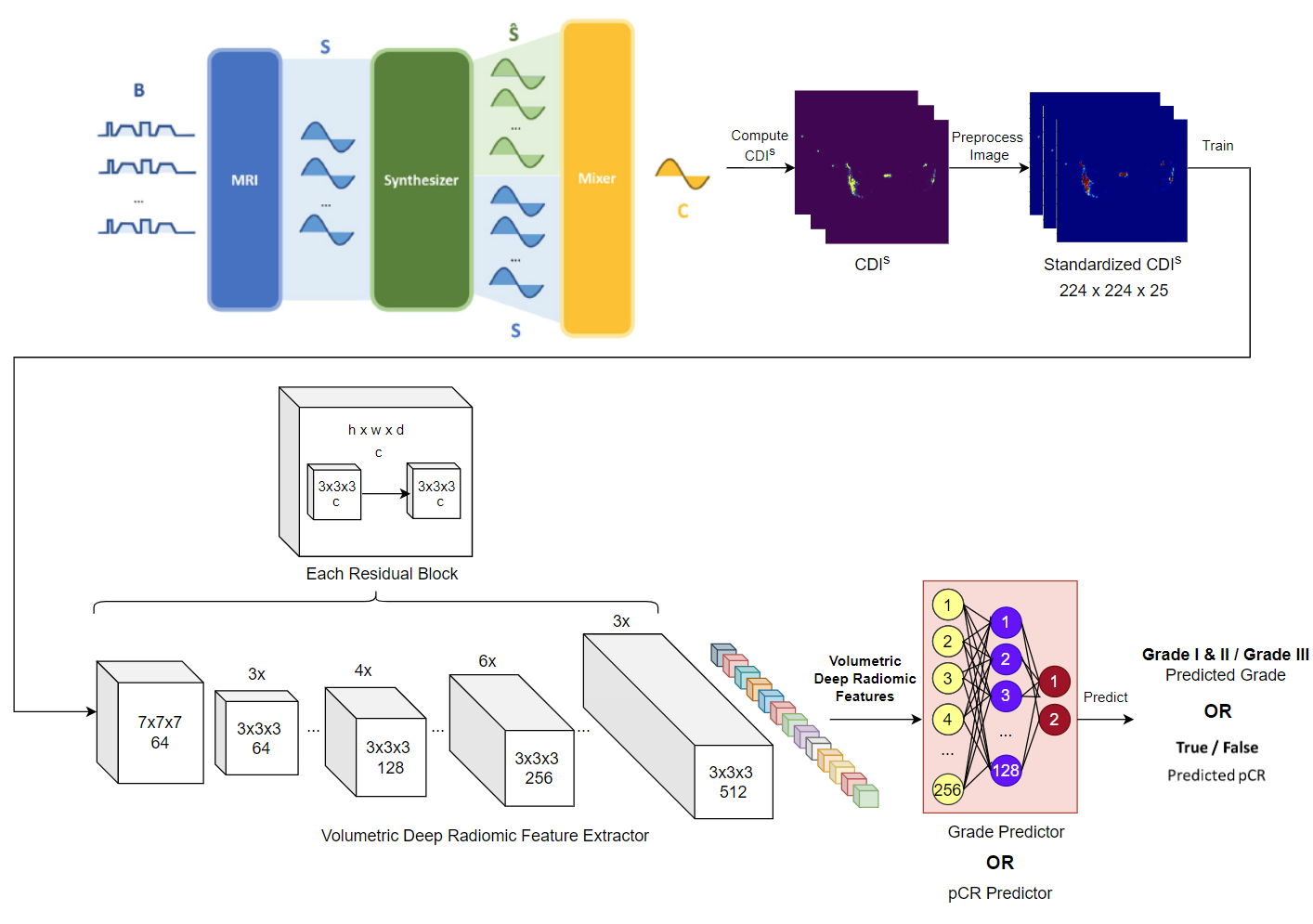}
  \caption{Clinical support workflow for grade (and pCR) prediction using volumetric deep radiomic features from synthetic correlated diffusion imaging (CDI\textsuperscript{s}).}
  \label{fig:workflow-figure}
  \vspace{-1em}
\end{figure}
To facilitate for the investigation into the efficacy of volumetric deep radiomic features from CDI\textsuperscript{s} for the breast cancer tasks, in this study we introduce a clinical support workflow shown in Figure~\ref{fig:workflow-figure}.  More specifically, CDI\textsuperscript{s} acquisitions are first performed on a given patient, which involves the acquisition of multiple native DWI signals with different b-values, passing these signals into a signal synthesizer to produce synthetic signals and then mixing the native and synthetic signals together to obtain a final signal (CDI\textsuperscript{s})~\cite{prostate-cdis}. The CDI\textsuperscript{s} acquisitions are then standardized into 224$\times$224$\times$25 volumetric data cubes to achieve dimensionality consistency for machine learning purposes. Next, motivated by the advances in deep learning as well as the volumetric nature of CDI\textsuperscript{s} data, a 34-layer volumetric residual convolutional neural network architecture was constructed and leveraged to learn volumetric deep radiomic features from the standardized volumetric data cubes. The aim with leveraging volumetric deep learning at this stage is to, rather than design hand-crafted radiomic features, directly learn volumetric deep radiomic features from patient data that characterizes the intrinsic properties of breast cancer tissue as captured by CDI\textsuperscript{s} that are relevant as it relates to cancer SBR grading and patient pCR to neoadjuvant chemotherapy after pre-treatment imaging. This volumetric neural network can then be used to produce deep radiomic features for each patient based on their CDI\textsuperscript{s} data cubes. Finally, a grading (or pCR) predictor comprising of a fully-connected neural network architecture is then learnt based on the extracted deep radiomic feature and SBR grading (or patient post-treatment pCR) data, and subsequently used to predict patient SBR grade (or pCR post-treatment to neoadjuvant chemotherapy).  

To evaluate the efficacy of the proposed approach, we conducted leave-one-out cross-validation (LOOCV) on the patient cohort with accuracy being the performance metric of interest. For comparison consistency, a separate volumetric deep radiomic feature extractor and grade (or pCR) predictor (with the same network architectures as for CDI\textsuperscript{s} as described in the Deep Learning Method Setup section) was used to learn a set of volumetric deep radiomic features from each gold-standard MRI modality (DWI, T2w, and ADC). 

As seen in Table~\ref{sbr-grade-dist}, there is an uneven distribution of patients between the three grades and hence, SBR grade I and II were combined into one category. 

\begin{table}[h]
    \caption{SBR grade distribution in the patient cohort.}
    \centering
    \NiceMatrixOptions{notes/para}
    \begin{NiceTabular}{l c}
        \toprule
        \RowStyle{\bfseries}
        SBR Grade & Number of Patients \\ \midrule
        Grade I (Low) & 5 \\
        Grade II (Intermediate) & 72 \\
        Grade III (High) & 175 \\ \bottomrule
    \end{NiceTabular}
    \label{sbr-grade-dist}
    \vspace{-1em}
\end{table}

\subsection{pCR Prediction via Volumetric Deep Radiomic Features}
Notably, we also leveraged DWI acquisitions in two different ways: 1) individual sets of features are learnt from DWI acquisitions of each b-value (b=0, 100, 600, 800), and 2) an individual set of features are also learnt from the combined stack of DWI acquisitions (namely, the b-values are treated as another channel in the input).

\section{Results}
\subsection{Breast Cancer Grading (SBR Grade)}
\begin{figure}
  \centering
    \includegraphics[width=\textwidth]{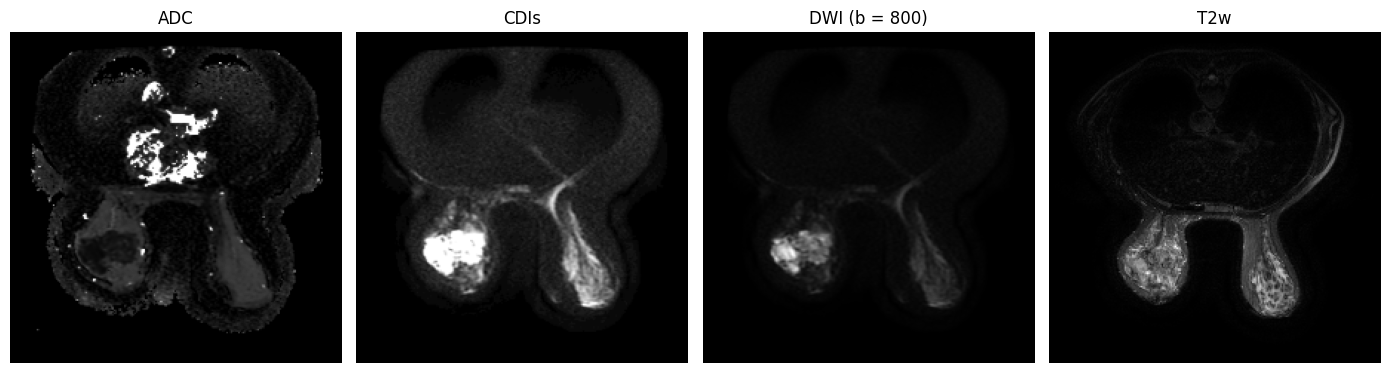}
  \caption{An example slice illustrating visual differences between ADC, CDI\textsuperscript{s}, DWI, and T2w at pre-treatment for a patient who has SBR Grade II (Intermediate). In this patient case, grade prediction was correct for CDI\textsuperscript{s} but not the other modalities.}
  \label{fig:grade-comparison-results}
\end{figure}

As seen in Table~\ref{grade-performance}, leveraging volumetric deep radiomic features for CDI\textsuperscript{s} achieves the highest grade predictive accuracy of 87.7\% with both sensitivity and specificity values over 80\%. Furthermore, CDI\textsuperscript{s} outperforms the gold-standard imaging modalities with an improvement of over 10\% on the next highest modality (T2w). With the highest gold-standard MRI modality only achieving a prediction accuracy of 76.59\%, over 10\% lower than CDI\textsuperscript{s}, the proposed approach with CDI\textsuperscript{s} can increase the grade prediction performance compared to gold-standard MRI modalities. An illustrative example highlighting the visual differences between the imaging modalities of ADC, CDI\textsuperscript{s}, DWI (b=800), and T2w for a patient case where grade prediction was correct for CDI\textsuperscript{s} but not the other modalities is shown Figure~\ref{fig:grade-comparison-results}.

\begin{table}[h]
    \caption{SBR grade prediction accuracy using LOOCV for different imaging modalities.}
    \vspace{-1em}
    \centering
    \NiceMatrixOptions{notes/para}
    \begin{NiceTabular}{lccc}
        \toprule
        \RowStyle{\bfseries}
        Modality & \multicolumn{1}{l}{Accuracy} & \multicolumn{1}{l}{Sensitivity} & \multicolumn{1}{l}{Specificity} \\
        \midrule
        \textbf{CDIs} & \textbf{87.70\%} & \textbf{90.29\%} & \textbf{81.82\%} \\
        T2w & 76.59\% & 99.43\% & 24.68\% \\
        ADC & 69.44\% & 100.00\% & 0.00\% \\
        DWI & 69.44\% & 95.43\% & 10.39\% \\ \bottomrule
    \end{NiceTabular}
    \label{grade-performance}
    \vspace{-2em}
\end{table}

\subsection{pCR Prediction via Volumetric Deep Radiomic Features}
\begin{figure}
\vspace{-0.5em}
  \centering
    \includegraphics[width=\textwidth]{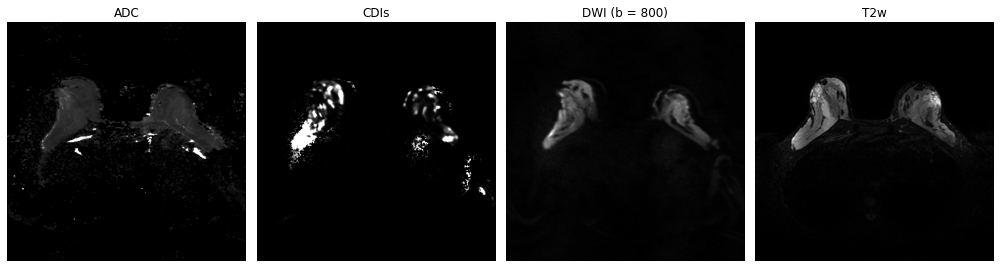}
  \caption{An example slice illustrating visual differences between ADC, CDI\textsuperscript{s}, DWI, and T2w before neoadjuvant chemotherapy for a patient who experienced pCR. In this patient case, pCR prediction was correct for CDI\textsuperscript{s} and DWI (b=800) but not the other modalities.}
  \label{fig:comparison-horizontal}
  \vspace{-1em}
\end{figure}

As seen in Table~\ref{loocv-results}, with the exception of ADC, leveraging volumetric deep radiomic features from each of the imaging modalities achieved pCR predictive accuracy over 80\% with the highest accuracy obtained from the CDI\textsuperscript{s} imaging modality. The volumetric deep radiomic features learnt using CDI\textsuperscript{s} enabled a pCR prediction accuracy of 87.75\%, which is over 3\% above the next best gold-standard MRI modality (i.e., DWI (b=800)). An illustrative example highlighting the visual differences between the imaging modalities of ADC, CDI\textsuperscript{s}, DWI (b=800), and T2w for a patient case where pCR prediction was correct for CDI\textsuperscript{s} and DWI (b=800) but not the other modalities is shown Figure~\ref{fig:comparison-horizontal}.

\begin{table}[h]
    \caption{pCR prediction accuracy using LOOCV for different imaging modalities.}
    \centering
    \NiceMatrixOptions{notes/para}
    \begin{NiceTabular}{l c}
        \toprule
        \RowStyle{\bfseries}
        Imaging Modality & Accuracy (\%) \\ \midrule
        \textbf{CDI\textsuperscript{s}} & \textbf{87.75} \\ 
         ADC & 79.84 \\
         T2w & 83.79 \\
         DWI (b=0, 100, 600, 800) & 84.19 \\
         DWI (b=0) & 84.19 \\
         DWI (b=100) & 82.21 \\
         DWI (b=600) & 84.19 \\
         DWI (b=800) & 84.58 \\ \bottomrule
    \end{NiceTabular}
    \label{loocv-results}
\end{table}

\section{Conclusion}
In this paper, we investigate enhancing clinical support for breast cancer with deep learning models using a newly introduced magnetic resonance imaging (MRI) modality called synthetic correlated diffusion imaging (CDI\textsuperscript{s}). Specifically, we leverage patients in the ACRIN-6698 study for the predictive tasks of breast cancer SBR grading and pCR prediction after neoadjuvant chemotherapy. Evaluated against current gold-standard imaging modalities, synthetic correlated diffusion imaging shows to enhance clinical support for breast cancer when paired with deep learning models, achieving a prediction accuracy of 87.70\% and 87.75\% for grading and pathologic complete response respectively. Given the promising results and higher performance of using CDI\textsuperscript{s} over the current gold-standard MRI images, future work involves expanding the study with a larger patient cohort to further validate our findings and leveraging improved CDI\textsuperscript{s} coefficient optimization to improve prediction performance.

\subsubsection{Prospect of application:}
The two models can be applied for the clinical tasks of breast cancer grading and pCR prediction respectively. It is envisioned that these models will be used by pathologists as an aid and to help identify patterns for more efficient breast cancer grading and effective treatment planning. These models should not be deployed standalone, rather as a tool for clinicians.

\bibliographystyle{splncs04}
\bibliography{bibliography}

\begin{thebibliography}{10}
\providecommand{\url}[1]{\texttt{#1}}
\providecommand{\urlprefix}{URL }
\providecommand{\doi}[1]{https://doi.org/#1}

\bibitem{sbr-useful}
Amat, S., et~al.: Scarff-bloom-richardson (sbr) grading: a pleiotropic marker
  of chemosensitivity in invasive ductal breast carcinomas treated by
  neoadjuvant chemotherapy. Int J Oncology  \textbf{4},  791--6 (2002),
  \url{https://link.springer.com/article/10.1007/s00428-021-03141-2}

\bibitem{american-breast-cancer}
{American Cancer Society}: Breast cancer.
  \url{https://www.cancer.org/cancer/breast-cancer} (2022), accessed:
  2023-08-03

\bibitem{comp-tumors}
Burnside, E.S., et~al.: Using computer-extracted image phenotypes from tumors
  on breast magnetic resonance imaging to predict breast cancer pathologic
  stage. ACS Journals  \textbf{122}(5),  748--757 (2016),
  \url{https://acsjournals.onlinelibrary.wiley.com/doi/full/10.1002/cncr.29791}

\bibitem{cancer-stats}
{Cancer.NET}: Breast cancer - statistics.
  \url{https://www.cancer.net/cancer-types/breast-cancer/statistics} (2023),
  accessed : 2023-06-10

\bibitem{acrin6698-data-4}
Clark, K., et~al.: The cancer imaging archive (tcia): Maintaining and operating
  a public information repository. Journal of Digital Imaging  \textbf{26}(6),
  1045--1057 (2013)

\bibitem{radiomics}
Conti, A., et~al.: Radiomics in breast cancer classification and prediction.
  Seminars in Cancer Biology  \textbf{72},  238--250 (2021).
  \doi{https://doi.org/10.1016/j.semcancer.2020.04.002},
  \url{https://www.sciencedirect.com/science/article/pii/S1044579X20300833},
  precision Medicine in Breast Cancer

\bibitem{image-analysis}
Couture, H.D., et~al.: Image analysis with deep learning to predict breast
  cancer grade, er status, histologic subtype, and intrinsic subtypes. npj
  Breast Cancer  \textbf{4}(30) (2018),
  \url{https://www.nature.com/articles/s41523-018-0079-1}

\bibitem{survival-bc-patients}
Dimitrakakis, C., Keramopoulos, A.: Survival in primary inoperable breast
  cancer patients. European Journal of Gynaecological Oncology  \textbf{25}(3),
   367--372 (2004)

\bibitem{grading-future}
van Dooijeweert, C., et~al.: Grading of invasive breast carcinoma: the way
  forward. Virchows Archiv  \textbf{480},  33--43 (2022),
  \url{https://link.springer.com/article/10.1007/s00428-021-03141-2}

\bibitem{prediction-pcr-3dcnn}
Duanmu, H., et~al.: Prediction of pathological complete response to neoadjuvant
  chemotherapy in breast cancer using deep learning with integrative imaging,
  molecular and demographic data. In: Martel, A.L., et~al. (eds.) Medical Image
  Computing and Computer Assisted Intervention – MICCAI 2020. pp. 242--252.
  MICCAI, Springer International Publishing, Cham (2020)

\bibitem{dce}
Fan, M., et~al.: Joint prediction of breast cancer histological grade and ki-67
  expression level based on dce-mri and dwi radiomics. IEEE Journal of
  Biomedical and Health Informatics  \textbf{24}(6),  1632--1642 (2020),
  \url{https://ieeexplore.ieee.org/abstract/document/8915701}

\bibitem{nac-bc-r}
Fangberget, A., et~al.: Neoadjuvant chemotherapy in breast cancer-response
  evaluation and prediction of response to treatment using dynamic
  contrast-enhanced and diffusion-weighted mr imaging. European Radiology
  \textbf{21},  1188--1199 (2011)

\bibitem{elastography}
Grajo, J.R., Barr, R.G.: Strain elastography for prediction of breast cancer
  tumor grades. Journal of Ultrasound in Medicine  \textbf{35}(1) (2014),
  \url{https://onlinelibrary.wiley.com/doi/abs/10.7863/ultra.33.1.129}

\bibitem{predicting-post-neoadjuvant}
Ha, R., et~al.: Predicting post neoadjuvant axillary response using a novel
  convolutional neural network algorithm. Annals of Surgical Oncology
  \textbf{25},  3037--3043 (2018)

\bibitem{3d-cnn-bc-detect}
Haq, A.U., et~al.: 3dcnn: Three-layers deep convolutional neural network
  architecture for breast cancer detection using clinical image data. In: 17th
  International Computer Conference on Wavelet Active Media Technology and
  Information Processing (ICCWAMTIP). ICCWAMTIP, IEEE, Chengdu, China (2020)

\bibitem{pcr-bc}
{Jamie DePolo}: Pathologic complete response to targeted therapy before surgery
  linked to better survival for early-stage her2-positive breast cancer.
  \url{https://www.breastcancer.org/research-news/neoadjuvant-pcr-linked-to-better-survival}
  (2020), accessed: 2022-06-02

\bibitem{VM208346402021.72720}
Kawahara, D., Nagata, Y.: T1-weighted and t2-weighted mri image synthesis with
  convolutional generative adversarial networks. Reports of Practical Oncology
  and Radiotherapy  \textbf{26}(1),  35 -- 42 (2021).
  \doi{10.5603/RPOR.a2021.0005},
  \url{https://journals.viamedica.pl/rpor/article/view/RPOR.a2021.0005}

\bibitem{cost-effectiveness-nac}
Kunst, N., et~al.: Cost-effectiveness of neoadjuvant-adjuvant treatment
  strategies for women with erbb2 (her2)–positive breast cancer. JAMA Network
  Open  \textbf{3}(11) (2020)

\bibitem{biomarkers-predicting}
Li, X.B., et~al.: Biomarkers predicting pathologic complete response to
  neoadjuvant chemotherapy in breast cancer. American Journal of Clinical
  Pathology  \textbf{145}(6),  871--878 (2016)

\bibitem{MARTINEZHERAS2021490}
Martinez-Heras, E., et~al.: Diffusion-weighted imaging: Recent advances and
  applications. Seminars in Ultrasound, CT and MRI  \textbf{42}(5),  490--506
  (2021). \doi{https://doi.org/10.1053/j.sult.2021.07.006},
  \url{https://www.sciencedirect.com/science/article/pii/S0887217121000846},
  advances in Neuroradiology I

\bibitem{nac-human-recommendation}
Masood, S.: Neoadjuvant chemotherapy in breast cancers. Womens Health
  \textbf{12}(5),  480--491 (2016)

\bibitem{ml-qCT}
Moghadas-Dastjerdi, H., et~al.: Machine learning-based a priori chemotherapy
  response prediction in breast cancer patients using textural ct biomarkers.
  In: 42nd Annual International Conference of the IEEE Engineering in Medicine
  \& Biology Society (EMBC). EMBC, IEEE, Montreal, QC, Canada (2020)

\bibitem{extended-apriori-prediction}
Moghadas-Dastjerdi, H., et~al.: A priori prediction of tumour response to
  neoadjuvant chemotherapy in breast cancer patients using quantitative ct and
  machine learning. Scientific Reports  \textbf{10},  1188--1199 (2020)

\bibitem{acrin6698-data-2}
Newitt, D.C., et~al.: Test–retest repeatability and reproducibility of adc
  measures by breast dwi: Results from the acrin 6698 trial. Journal of
  Magnetic Resonance Imaging  \textbf{49}(6),  1617--1628 (2018)

\bibitem{acrin6698-data-3}
Newitt, D.C., et~al.: Acrin 6698/i-spy2 breast dwi [data set]. The Cancer
  Imaging Archive  (2021)

\bibitem{acrin6698-data-1}
Partridge, S.C., et~al.: Diffusion-weighted mri findings predict pathologic
  response in neoadjuvant treatment of breast cancer: The acrin 6698
  multicenter trial. Radiology  \textbf{289}(3),  618--627 (2018)

\bibitem{human-judgement-problem}
Redelmeier, D.A., et~al.: Problems for clinical judgement: introducing
  cognitive psychology as one more basic science. Canadian Medical Association
  Journal  \textbf{164}(3),  358--360 (2001)

\bibitem{prediction-dwi-mrs}
Shin, H.J., et~al.: Prediction of pathologic response to neoadjuvant
  chemotherapy in patients with breast cancer using diffusion-weighted imaging
  and mrs. NMR Biomedical  \textbf{25}(12),  1349--1359 (2012)

\bibitem{dwi-prediction}
Surov, A., et~al.: Can diffusion-weighted imaging predict tumor grade and
  expression of ki-67 in breast cancer? a multicenter analysis. Breast Cancer
  Research  \textbf{20}(58) (2018),
  \url{https://breast-cancer-research.biomedcentral.com/articles/10.1186/s13058-018-0991-1}

\bibitem{https://doi.org/10.48550/arxiv.1606.05718}
Wang, D., et~al.: Deep learning for identifying metastatic breast cancer
  (2016). \doi{10.48550/ARXIV.1606.05718},
  \url{https://arxiv.org/abs/1606.05718}

\bibitem{prostate-cdis}
Wong, A., et~al.: Synthetic correlated diffusion imaging hyperintensity
  delineates clinically significant prostate cancer. Scientific Reports
  \textbf{12}(3376) (2022), \url{https://doi.org/10.1038/s41598-022-06872-7}

\end{thebibliography}

\end{document}